# ECG Feature Extraction Techniques - A Survey Approach

S. Karpagachelvi,
Doctoral Research Scholar,
Mother Teresa Women's University,
Kodaikanal, Tamilnadu, India.
email : karpagachelvis@yahoo.com

Dr. M. Arthanari, Prof. & Head,
Dept. of Computer Science and Engineering,
Tejaa Shakthi Institute of Technology for Women,
Coimbatore- 641 659, Tamilnadu, India.
email: arthanarimsvc@gmail.com

M. Sivakumar,
Doctoral Research Scholar,
Anna University – Coimbatore,
Tamilnadu, India
email : sivala@gmail.com

*Abstract*—ECG Feature Extraction plays a significant role in diagnosing most of the cardiac diseases. One cardiac cycle in an ECG signal consists of the P-QRS-T waves. This feature extraction scheme determines the amplitudes and intervals in the ECG signal for subsequent analysis. The amplitudes and intervals value of P-QRS-T segment determines the functioning of heart of every human. Recently, numerous research and techniques have been developed for analyzing the ECG signal. The proposed schemes were mostly based on Fuzzy Logic Methods, Artificial Neural Networks (ANN), Genetic Algorithm (GA), Support Vector Machines (SVM), and other Signal Analysis techniques. All these techniques and algorithms have their advantages and limitations. This proposed paper discusses various techniques and transformations proposed earlier in literature for extracting feature from an ECG signal. In addition this paper also provides a comparative study of various methods proposed by researchers in extracting the feature from ECG signal.

## I. INTRODUCTION

The investigation of the ECG has been extensively used for diagnosing many cardiac diseases. The ECG is a realistic record of the direction and magnitude of the electrical commotion that is generated by depolarization and re-polarization of the atria and ventricles. One cardiac cycle in an ECG signal consists of the P-QRS-T waves. Figure 1 shows a sample ECG signal. The majority of the clinically useful information in the ECG is originated in the intervals and amplitudes defined by its features (characteristic wave peaks and time durations). The improvement of precise and rapid methods for automatic ECG feature extraction is of chief importance, particularly for the examination of long recordings [1].

The ECG feature extraction system provides fundamental features (amplitudes and intervals) to be used in subsequent automatic analysis. In recent times, a number of techniques have been proposed to detect these features [2] [3] [4]. The previously proposed method of ECG signal analysis was based on time domain method. But this is not always adequate to study all the features of ECG signals. Therefore the frequency representation of a signal is required. The deviations in the normal electrical patterns indicate various cardiac disorders. Cardiac cells, in the normal state are electrically polarized [5].

ECG is essentially responsible for patient monitoring and diagnosis. The extracted feature from the ECG signal plays a vital in diagnosing the cardiac disease. The development of accurate and quick methods for automatic ECG feature extraction is of major importance. Therefore it is necessary that the feature extraction system performs accurately. The purpose of feature extraction is to find as few properties as possible within ECG signal that would allow successful abnormality detection and efficient prognosis.

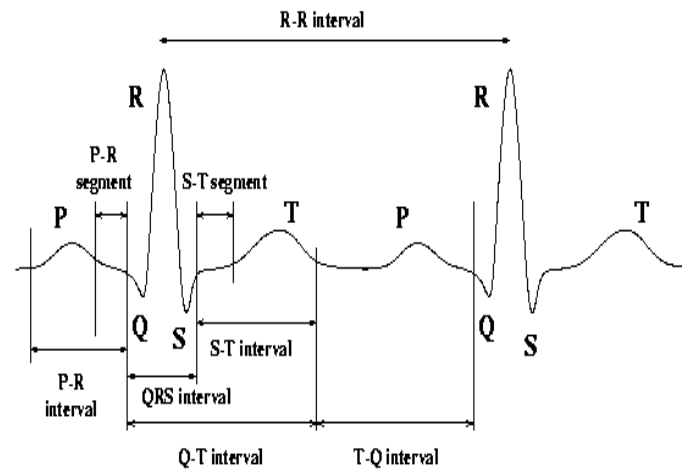

Figure.1 A Sample ECG Signal showing P-QRS-T Wave

In recent year, several research and algorithm have been developed for the exertion of analyzing and classifying the ECG signal. The classifying method which have been proposed during the last decade and under evaluation includes digital signal analysis, Fuzzy Logic methods, Artificial Neural Network, Hidden Markov Model, Genetic Algorithm, Support Vector Machines, Self-Organizing Map, Bayesian and other method with each approach exhibiting its own advantages and disadvantages. This paper provides an over view on various techniques and transformations used for extracting the feature from ECG signal. In addition the future enhancement gives a general idea for improvement and development of the feature extraction techniques.

The remainder of this paper is structured as follows. Section 2 discusses the related work that was earlier proposed in literature for ECG feature extraction. Section 3 gives a general idea of further improvements of the earlier approaches in ECG feature detection, and Section 4 concludes the paper with fewer discussions.

## II. LITERATURE REVIEW

ECG feature extraction has been studied from early time and lots of advanced techniques as well as transformations have been proposed for accurate and fast ECG feature extraction. This section of the paper discusses various techniques and transformations proposed earlier in literature for extracting feature from ECG.

Zhao et al. [6] proposed a feature extraction method using wavelet transform and support vector machines. The paper presented a new approach to the feature extraction for reliable heart rhythm recognition. The proposed system of classification is comprised of three components including data preprocessing, feature extraction and classification of ECG signals. Two diverse feature extraction methods are applied together to achieve the feature vector of ECG data. The wavelet transform is used to extract the coefficients of the transform as the features of each ECG segment. Concurrently, autoregressive modeling (AR) is also applied to get hold of the temporal structures of ECG waveforms. Then at last the support vector machine (SVM) with Gaussian kernel is used to classify different ECG heart rhythm. The results of computer simulations provided to determine the performance of the proposed approach reached the overall accuracy of 99.68%.

A novel approach for ECG feature extraction was put forth by Castro et al. in [7]. Their proposed paper present an algorithm, based on the wavelet transform, for feature extraction from an electrocardiograph (ECG) signal and recognition of abnormal heartbeats. Since wavelet transforms can be localized both in the frequency and time domains. They developed a method for choosing an optimal mother wavelet from a set of orthogonal and bi-orthogonal wavelet filter bank by means of the best correlation with the ECG signal. The foremost step of their approach is to denoise (remove noise) the ECG signal by a soft or hard threshold with limitation of 99.99 reconstructs ability and then each PQRST cycle is decomposed into a coefficients vector by the optimal wavelet function. The coefficients, approximations of the last scale level and the details of the all levels, are used for the ECG analyzed. They divided the coefficients of each cycle into three segments that are related to P-wave, QRS complex, and T-wave. The summation of the values from these segments provided the feature vectors of single cycles.

Mahmoodabadi et al. in [1] described an approach for ECG feature extraction which utilizes Daubechies Wavelets transform. They had developed and evaluated an electrocardiogram (ECG) feature extraction system based on the multi-resolution wavelet transform. The ECG signals from Modified Lead II (MLII) were chosen for processing. The wavelet filter with scaling function further intimately similar to the shape of the ECG signal achieved better detection. The foremost step of their approach was to de-noise the ECG signal by removing the equivalent wavelet coefficients at higher scales. Then, QRS complexes are detected and each one complex is used to trace the peaks of the individual waves, including onsets and offsets of the P and T waves which are present in one cardiac cycle. Their experimental results revealed that their proposed approach for ECG feature extraction achieved sensitivity of 99.18% and a positive predictivity of 98%.

A Mathematical morphology for ECG feature extraction was proposed by Tadejko and Rakowski in [8]. The primary focus of their work is to evaluate the classification performance of an automatic classifier of the electrocardiogram (ECG) for the detection abnormal beats with new concept of feature extraction stage. The obtained feature sets were based on ECG morphology and RR-intervals. Configuration adopted a well known Kohonen self-organizing maps (SOM) for examination of signal features and clustering. A classifier was developed with SOM and learning vector quantization (LVQ) algorithms using the data from the records recommended by ANSI/AAMI EC57 standard. In addition their work compares two strategies for classification of annotated QRS complexes: based on original ECG morphology features and proposed new approach - based on preprocessed ECG morphology features. The mathematical morphology filtering is used for the preprocessing of ECG signal.

Sufi et al. in [9] formulated a new ECG obfuscation method for feature extraction and corruption detection. They present a new ECG obfuscation method, which uses cross correlation based template matching approach to distinguish all ECG features followed by corruption of those features with added noises. It is extremely difficult to reconstruct the obfuscated features without the knowledge of the templates used for feature matching and the noise. Therefore, they considered three templates and three noises for P wave, QRS Complex and T wave comprise the key, which is only 0.4%-0.9% of the original ECG file size. The key distribution among the authorized doctors is efficient and fast because of its small size. To conclude, the experiments carried on with unimaginably high number of noise combinations the security strength of the presented method was very high.

Saxena et al in [10] described an approach for effective feature extraction form ECG signals. Their paper deals with an competent composite method which has been developed for data compression, signal retrieval and feature extraction of ECG signals. After signal retrieval from the compressed data, it has been found that the network not only compresses the data, but also improves the quality of retrieved ECG signal with respect to elimination of high-frequency interference present in the original signal. With the implementation of artificial neural network (ANN) the compression ratio increases as the number of ECG cycle increases. Moreover the features extracted by amplitude, slope and duration criteria from the retrieved signal match with the features of the original signal. Their experimental results at every stage are steady and consistent and prove beyond doubt that the composite method can be used for efficient data management and feature extraction of ECG signals in many real-time applications.

A feature extraction method using Discrete Wavelet Transform (DWT) was proposed by Emran et al. in [11]. They used a discrete wavelet transform (DWT) to extract the relevant information from the ECG input data in order to perform the classification task. Their proposed work includes the following modules data acquisition, pre-processing beat detection, feature extraction and classification. In the feature extraction module the Wavelet Transform (DWT) is designed to address the problem of non-stationary ECG signals. It was derived from a single generating function called the mother wavelet by translation and dilation operations. Using DWT in feature extraction may lead to an optimal frequency resolution in all frequency ranges as it has a varying window size, broad at lower frequencies, and narrow at higher frequencies. The DWT characterization will deliver the stable features to the morphology variations of the ECG waveforms.

Tayel and Bouridy together in [12] put forth a technique for ECG image classification by extracting their feature using wavelet transformation and neural networks. Features are extracted from wavelet decomposition of the ECG images intensity. The obtained ECG features are then further processed using artificial neural networks. The features are: mean, median, maximum, minimum, range, standard deviation, variance, and mean absolute deviation. The introduced ANN was trained by the main features of the 63 ECG images of different diseases. The test results showed that the classification accuracy of the introduced classifier was up to 92%. The extracted features of the ECG signal using wavelet decomposition was effectively utilized by ANN in producing the classification accuracy of 92%.

Alan and Nikola in [13] proposed chaos theory that can be successfully applied to ECG feature extraction. They also discussed numerous chaos methods, including phase space and attractors, correlation dimension, spatial filling index, central tendency measure and approximate entropy. They created a new feature extraction environment called ECG chaos extractor to apply the above mentioned chaos methods. A new semi-automatic program for ECG feature extraction has been implemented and is presented in this article. Graphical interface is used to specify ECG files employed in the extraction procedure as well as for method selection and results saving. The program extracts features from ECG files.

An algorithm was presented by Chouhan and Mehta in [14] for detection of QRS complexities. The recognition of QRS-complexes forms the origin for more or less all automated ECG analysis algorithms. The presented algorithm utilizes a modified definition of slope, of ECG signal, as the feature for detection of QRS. A succession of transformations of the filtered and baseline drift corrected ECG signal is used for mining of a new modified slope-feature. In the presented algorithm, filtering procedure based on moving averages [15] provides smooth spike-free ECG signal, which is appropriate for slope feature extraction. The foremost step is to extort slope feature from the filtered and drift corrected ECG signal, by processing and transforming it, in such a way that the extracted feature signal is significantly enhanced in QRS region and suppressed in non-QRS region. The proposed method has detection rate and positive predictivity of 98.56% and 99.18% respectively.

Xu et al. in [16] described an algorithm using Slope Vector Waveform (SVW) for ECG QRS complex detection and RR interval evaluation. In their proposed method variable stage differentiation is used to achieve the desired slope vectors for feature extraction, and the non-linear amplification is used to get better of the signal-to-noise ratio. The method allows for a fast and accurate search of the R location, QRS complex duration, and RR interval and yields excellent ECG feature extraction results. In order to get QRS durations, the feature extraction rules are needed.

A method for automatic extraction of both time interval and morphological features, from the Electrocardiogram (ECG) to classify ECGs into normal and arrhythmic was described by Alexakis et al. in [17]. The method utilized the combination of artificial neural networks (ANN) and Linear Discriminant Analysis (LDA) techniques for feature extraction. Five ECG features namely RR, RTc, T wave amplitude, T wave skewness, and T wave kurtosis were used in their method. These features are obtained with the assistance of automatic algorithms. The onset and end of the T wave were detected using the tangent method. The three feature combinations used had very analogous performance when considering the average performance metrics.

A modified combined wavelet transforms technique was developed by Saxena et al. in [18]. The technique has been developed to analyze multi lead electrocardiogram signals for cardiac disease diagnostics. Two wavelets have been used, i.e. a quadratic spline wavelet (QSWT) for QRS detection and the Daubechies six coefficient (DU6) wavelet for P and T detection. A procedure has been evolved using electrocardiogram parameters with a point scoring system for diagnosis of various cardiac diseases. The consistency and reliability of the identified and measured parameters were confirmed when both the diagnostic criteria gave the same results. Table 1 shows the comparison of different ECG signal feature extraction techniques.

A robust ECG feature extraction scheme was put forth by Olvera in [19]. The proposed method utilizes a matched filter to detect different signal features on a human heart electrocardiogram signal. The detection of the ST segment, which is a precursor of possible cardiac problems, was more difficult to extract using the matched filter due to noise and amplitude variability. By improving on the methods used; using a different form of the matched filter and better threshold detection, the matched filter ECG feature extraction could be made more successful. The detection of different features in the ECG waveform was much harder than anticipated but it was not due to the implementation of the matched filter. The more complex part was creating the revealing method to remove the feature of interest in each ECG signal.

Jen et al. in [20] formulated an approach using neural networks for determining the features of ECG signal. They presented an integrated system for ECG diagnosis. The integrated system comprised of cepstrum coefficient method for feature extraction from long-term ECG signals and artificial neural network (ANN) models for the classification. Utilizing the proposed method, one can identify the characteristics hiding inside an ECG signal and then classify the signal as well as diagnose the abnormalities. To explore the performance of the proposed method various types of ECG data from the MIT/BIH database were used for verification. The experimental results showed that the accuracy of diagnosing cardiac disease was above 97.5%. In addition the proposed method successfully extracted the corresponding feature vectors, distinguished the difference and classified ECG signals.

Correlation analysis for abnormal ECG signal feature extraction was explained by Ramli and Ahmad in [21]. Their proposed work investigated the technique to extract the important features from the 12 lead system (electrocardiogram) ECG signals. They chose II for their entire analysis due to its representative characteristics for identifying the common heart diseases. The analysis technique chosen is the cross-correlation analysis. Cross-correlation analysis measures the similarity between the two signals and extracts the information present in the signals. Their test results suggested that the proposed technique could effectively extract features, which differentiate between the types of heart diseases analyzed and also for normal heart signal.

Ubeyli et al. in [22] described an approach for feature extraction from ECG signal. They developed an automated diagnostic systems employing dissimilar and amalgamated features for electrocardiogram (ECG) signals were analyzed and their accuracies were determined. The classification accuracies of mixture of experts (ME) trained on composite features and modified mixture of experts (MME) trained on diverse features were also compared in their work. The inputs of these automated diagnostic systems were composed of diverse or composite features and these were chosen based on the network structures. The achieved accuracy rates of their proposed approach were higher than that of the ME trained on composite features.

Fatemian et al. [25] proposed an approach for ECG feature extraction. They suggested a new wavelet based framework for automatic analysis of single lead electrocardiogram (ECG) for application in human recognition. Their system utilized a robust preprocessing stage, which enables it to handle noise and outliers. This facilitates it to be directly applied on the raw ECG signal. In addition the proposed system is capable of managing ECGs regardless of the heart rate (HR) which renders making presumptions on the individual's stress level unnecessary. The substantial reduction of the template gallery size decreases the storage requirements of the system appreciably. Additionally, the categorization process is speeded up by eliminating the need for dimensionality reduction techniques such as PCA or LDA. Their experimental results revealed the fact that the proposed technique out performed other conventional methods of ECG feature extraction.

III. FUTURE ENHANCEMENT

The electrocardiogram (ECG) is a noninvasive and the record of variation of the bio-potential signal of the human heartbeats. The ECG detection which shows the information of the heart and cardiovascular condition is essential to enhance the patient living quality and appropriate treatment. The ECG features can be extracted in time domain [23] or in frequency domain [24]. The extracted feature from the ECG signal plays a vital in diagnosing the cardiac disease. The development of accurate and quick methods for automatic ECG feature extraction is of major importance. Some of the features extraction methods implemented in previous research includes Discrete Wavelet Transform, Karhunen-Loeve Transform, Hermitian Basis and other methods. Every method has its own advantages and limitations. The future work primarily focus on feature extraction from an ECG signal using more statistical data. In addition the future enhancement eye on utilizing different transformation technique that provides higher accuracy in feature extraction. The parameters that must be considered while developing an algorithm for feature extraction of an ECG signal are simplicity of the algorithm and the accuracy of the algorithm in providing the best results in feature extraction.

Table I. Comparison of Different Feature Extraction Techniques from an ECG Signal where H, M, L denotes High, Medium and Low respectively

| Approach | Simplicity | Accuracy | Predictivity |
|---|---|---|---|
| Zhao et al. | H | H | H |
| Mahmoodabadi et al. | M | H | H |
| Tadejko and Rakowski | L | M | M |
| Tayel and Bouridy | M | M | H |
| Jen et al. | H | H | H |
| Alexakis et al. | H | M | M |
| Ramli and Ahmad | M | M | M |
| Xu et al. | M | H | H |
| Olvera | H | M | M |
| Emran et al | H | M | L |

IV. CONCLUSION

The examination of the ECG has been comprehensively used for diagnosing many cardiac diseases. Various techniques and transformations have been proposed earlier in literature for extracting feature from ECG. This proposed paper provides an over view of various ECG feature extraction techniques and algorithms proposed in literature. The feature extraction technique or algorithm developed for ECG must be highly accurate and should ensure fast extraction of features from the ECG signal. This proposed paper also revealed a comparative table evaluating the performance of different algorithms that were proposed earlier for ECG signal feature extraction. The future work mainly concentrates on developing an algorithm for accurate and fast feature extraction. Moreover additional statistical data will be utilized for evaluating the performance of an algorithm in ECG signal feature detection. Improving the accuracy of diagnosing the cardiac disease at the earliest is necessary in the case of patient monitoring system. Therefore our future work also has an eye on improvement in diagnosing the cardiac disease.

## AUTHORS PROFILE

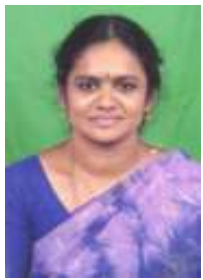

Karpagachelvi.S: She received the BSc degree in physics from Bharathiar University in 1993 and Masters in Computer Applications from Madras University in 1996. She has 12 years of teaching experience. She is currently a PhD student with the Department of Computer Science at Mother Teresa University.

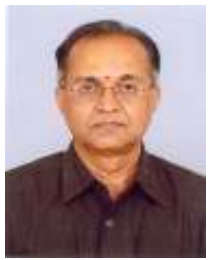

Dr.M.Arthanari: He has obtained Doctorate in Mathematics in Madras University in the year 1981. He has 35 years of teaching experience and 25 years of research experience. He has a Patent in Computer Science approved by Govt. of India.

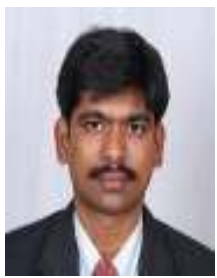

Sivakumar M : He has 10+ years of experience in the software industry including Oracle Corporation. He received his Bachelor degree in Physics and Masters in Computer Applications from the Bharathiar University, India. He holds patent for the invention in embedded technology. He is technically certified by various professional bodies like ITIL, IBM Rational Clearcase Administrator, OCP - Oracle Certified Professional 10g and ISTQB.